%% file: main.tex
\documentclass[letterpaper]{article} 
\usepackage{aaai24}  
\usepackage{times}  
\usepackage{helvet}  
\usepackage{courier}  
\usepackage[hyphens]{url}  
\usepackage{graphicx} 
\usepackage{natbib}  
\usepackage{caption} 
\usepackage{algorithm}
\usepackage{algorithmic}
\usepackage{times}
\usepackage{latexsym}
\usepackage{booktabs}
\usepackage{graphicx}
\usepackage{xcolor}
\usepackage{amsmath}
\usepackage{amsfonts}
\usepackage{cleveref}
\usepackage{subfigure}

\usepackage{newfloat}
\usepackage{listings}
\DeclareCaptionStyle{ruled}{labelfont=normalfont,labelsep=colon,strut=off} 
\floatstyle{ruled}
\newfloat{listing}{tb}{lst}{}
\floatname{listing}{Listing}
\newcommand{\wy}[1]{{\color{black}{#1}}}

\title{Deciphering Compatibility Relationships with Textual Descriptions via Extraction and Explanation}
\author{
    Yu Wang,
    Zexue He,
    Zhankui He,
    Hao Xu,
    Julian McAuley
}
\affiliations{
    University of California, San Diego\\
    \{yuw164, zehe, zhh004, hax019, jmcauley\}@ucsd.edu
}

\usepackage{bibentry}

\begin{document}

\maketitle

\begin{abstract}
Understanding and accurately explaining compatibility relationships between fashion items is a challenging problem in the burgeoning domain of AI-driven outfit recommendations. 
Present models, while making strides in this area, still occasionally fall short, offering explanations that can be elementary and repetitive. 
This work aims to address these shortcomings by introducing the Pair Fashion Explanation (PFE) dataset, a unique resource that has been curated to illuminate these compatibility relationships. Furthermore, we propose an innovative two-stage pipeline model that leverages this dataset.
This fine-tuning allows the model to generate explanations that convey the compatibility relationships between items. Our experiments showcase the model's potential in crafting descriptions that are knowledgeable, aligned with ground-truth matching correlations, and that produce understandable and informative descriptions, as assessed by both automatic metrics and human evaluation. Our code and data are released at \url{https://github.com/wangyu-ustc/PairFashionExplanation}.
\end{abstract}

\input{1_introduction}

\input{2_related_work}
\input{3_method}
\input{4_experiments}
\input{5_conclusion}

\bibliography{aaai24}
\appendix
\newpage
\input{7_appendix}

\end{document}

%% file: 1_introduction.tex
\vspace{-5pt}
\section{Introduction}
\label{sec:introduction}



Fashion and technology now intersect 
through outfit recommendation platforms, which provide a vast array of apparel and accessory options.
These platforms are prevalent, such as Amazon Fashion\footnote{\url{https://www.amazon.com/amazon-fashion/b?ie=UTF8&node=7141123011}} and Chictopia\footnote{\url{http://www.chictopia.com/}}. 
An essential component of these recommendation systems is explanation capabilities, which guide user decisions and enable system diagnosis. Additionally, as outfit recommendation becomes popular, the need for explanatory functionality extends beyond single items, emphasizing the importance of illustrating compatibility relationships between pairs of items. 
This presents the central challenge that our research seeks to address: \textbf{Given a pair of fashion items, how can we effectively uncover and articulate their intrinsic compatibility relationships?} In order to ensure these explanations are comprehensible to a broad spectrum of users, we prioritize generating explanations as coherent and naturally-phrased sentences.


Existing literature offers a multitude of models for outfit recommendations, which are designed to generate highly compatible outfit combinations~\citep{outfit_compatibility_prediction,POG}. However, these models often overlook the essential component of providing explanations for their predictions. Certain efforts have been made to incorporate explanatory features within prediction models~\citep{OutfitRecommendationwithComplementaryClothingMatching, VICTOR}, but these attempts frequently result in rudimentary and repetitive explanations, failing to accurately encapsulate the complex compatibility relationships between diverse pairs of clothing items~\citep{explainable_outfit_recommendation,kaicheng2021modeling}. This indicates an unresolved challenge in the field: the generation of nuanced and informative explanations that accurately reflect true compatibility relationships.

\begin{figure}
    \centering
    \includegraphics[width=\linewidth]{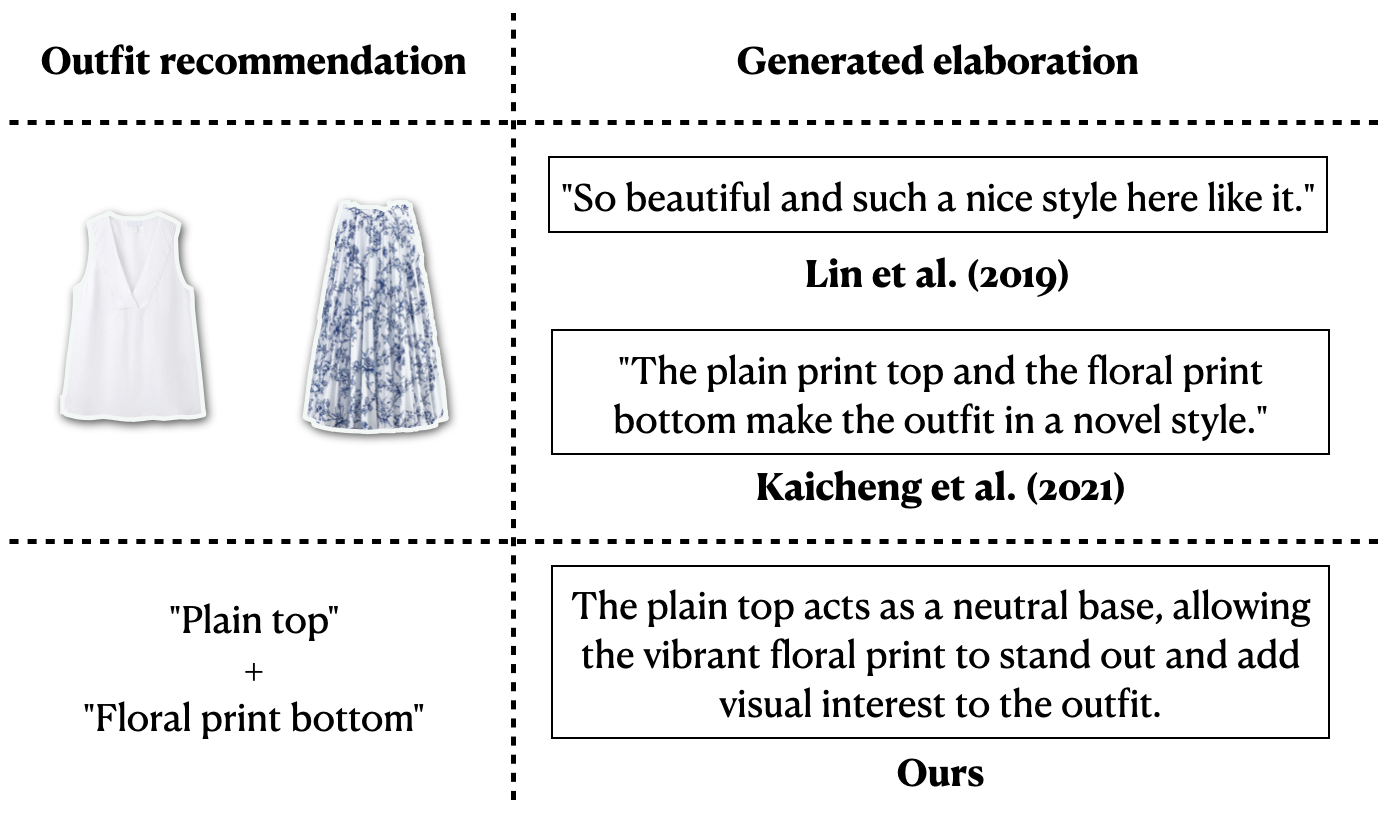}
    \caption{Our task is generating natural language descriptions to explain compatibility relationships between items.}
    \label{fig:introduction}
    \vspace*{-15pt}
\end{figure}

Recent advancements in large language models like ChatGPT and GPT4~\citep{openai2023gpt} have shown considerable potential across various tasks~\citep{chatgpteval}. However, their ability to generate expert-aligned explanations for clothing matching remains dubious, with a tendency to produce plausible yet inauthentic explanations that diverge from the true compatibility relationships.

To address this challenge, we present a 
novel solution combining a newly curated dataset and a two-stage pipeline model. 
Our \textbf{PFE} dataset comprises 6,407 
sentences.  Each sentence in the dataset serves as a distinct compatibility explanation for a pair of items, providing relevant entities and features. It offers insights into the explanations of clothing compatibility. By focusing on quality rather than quantity, our \textbf{PFE} dataset provides dense, high-quality data that can effectively support the intricate task of elucidating clothing compatibility relationships. 

Leveraging the PFE dataset, we further develop a two-stage pipeline that extracts important features from item descriptions and generates corresponding explanations. 
In the first stage, we use a large general public dataset collected from user interactions to train a feature extractor model. This model extracts important features from the given abundant features with the supervision of positive and negative pairs. 
The second stage utilizes our PFE dataset to fine-tune a large language model. With the explicit compatibility relationships demonstrated in the dataset, we can inject the domain-specific knowledge into the large language model, enabling 
the model to discover compatibility relationships between items. During inference, our proposed framework also employs a two-stage approach. We use the first-stage model to extract important features and make predictions. If the prediction is positive, the extracted features are templated as prompts for the second-stage model to generate explanations. 
This approach ensures that the generated explanations are more likely to make sense and extract the inherent relationships between items.
The proposed pipeline allows for a more nuanced and informative approach to generating explanations for item matching. 

Our contributions can be summarized as:
\begin{enumerate}
    \item We curate a novel dataset specifically designed for pair-matching explanations. To the best of our knowledge, this is the first dataset for pairwise explanations. This dataset helps us to fine-tune our model for better performance.
    \item We introduce a pipeline that utilizes the proposed dataset and other relevant datasets to extract important features and generate informative and diversified explanations. 
    \item Experimental results demonstrate that our model generates knowledgeable and aligned descriptions that are more accurate in terms of ground-truth matching correlations compared to existing methods. Meanwhile, our model produces more acceptable and informative descriptions according to human evaluation compared to other recommendation models and general-purpose language models.
\end{enumerate}

%% file: 2_related_work.tex
\section{Related Work}
\paragraph{Outfit Recommendation}
Within the realm of outfit recommendation, research efforts can broadly be categorized into several strands. A segment of these studies emphasizes recommending comprehensive sets of clothing items, treating outfits as integral units~\citep{POG, PORLA,heterogeneousGraphForOutfitRecommendation,DBLP:journals/tist/BanerjeeRSG20}. Others approach the problem from the perspective of compatibility predictions, generating compatibility scores for individual items within a partial outfit~\citep{OutfitTransformer,MOCM-MGL,AHGN,OutfitGAN}. Alternatively, there are methods that compute compatibility scores for pairs of items, instrumental in the construction of outfits or item retrieval~\citep{OutfitNet,balim2023diagnosing,li2019coherent,FashionComplementaryOutfitProductRetrieval,chen2022personalized}. Another line of research evaluates the overall compatibility score of an entire outfit~\citep{OutfitRecommendationwithComplementaryClothingMatching}. Certain studies have further explored the concept of incompatibility detection~\citep{VICTOR,balim2023diagnosing}, aiming to identify items within an outfit that mismatch. Although these works employ diverse methodologies, a common feature among them is the absence of predictive explanations.

\paragraph{Explanations for Outfit Recommendation}
While the bulk of outfit recommendation research concentrates on generating recommendations without explanations, a few works have 
tried
to incorporate justifications for their predictions. For instance, \citet{mo2023towards} predicts attributes of the missing item, utilizing the predicted attributes as explanations and indicators for item recommendation. In a similar vein, \citet{tangseng2020toward} 
detects incompatible items and provides basic reasons for incompatibility, such as mismatched texture or color. Beyond these rudimentary justifications, some research efforts seek to generate natural language explanations. \citet{kaicheng2021modeling} extracts item attributes and identifies those contributing to prediction, subsequently use them to construct explanatory sentences. \citet{explainable_outfit_recommendation} also generates textual explanations, although their scope is often limited and lacks detailed insight. In our study, we aim to address these limitations by generating detailed, diverse, and informative explanations for item pair compatibility within outfits.

%% file: 3_method.tex
\section{Methodology}
\subsection{Notation and Problem Formulation}
We start by defining an item pair $(i,j)$, where $i, j$ are the indices of the items.
Each item $i$ is associated with a distinct set of 
descriptive words, constructing a bag of words $t_i$,
which describe its color, style, material, and other properties. 
The features of the item pair $(i,j)$ are denoted as $(t_i,t_j)$. Additionally, we have the category information of the items, which is denoted as $(c_i,c_j)$ for the item pair $(i,j)$. 
For example, one data instance from FashionVC~\cite{neurostylist} for $t_i, t_j, c_i, c_j$ is ``Bamford Cropped cashmere cable-knit'', ``Vince Sequined georgette mini'', ``sweater'', ``skirt'', respectively. 

Given the item pair $(i, j)$ with their corresponding features and categories, our target is to generate the explanation for the pair, which is defined as $t_e$. To achieve this goal, we propose the extraction model $f_\theta$ and the generation model $g_\varphi$. The extraction model $f_\theta$ extracts the important features in $(t_i, t_j)$: 
\begin{equation}\label{eq:extraction_model}
    (r_i, r_j) = f_\theta(t_i, t_j)
\end{equation}
where $(r_i, r_j)$ are the important features in $(t_i, t_j)$ which contribute to whether items $(i, j)$ match or not. While $g_\varphi$ could generate the explanations to describe why $(i_1, i_2)$ is a matching pair: 
\begin{equation}\label{eq:generation_model}
    t_e = g_\varphi(c_i, r_i, c_j, r_j)
\end{equation}
Here $t_e$ is the explanation for the item pair $(i, j)$.

\subsection{Stage I: Extract important features}
\label{ssub: stage1_extract_important_features}
In this stage, we require a well-trained feature extraction model $f_\theta$ which needs supervision.
We use datasets of matching pairs, constructed as $\{i_{n_1}, i_{n_2}\}_{n=1}^{N}$. 
To obtain negative pairs, we sample items from the entire set and ensure exclusivity with the positive pairs to form $\{i_{n_1}, i_{n_2}\}_{n=N+1}^{2N}$. The positive and negative pairs provide the supervision to train $f_\theta$. However, since $f_\theta$ only performs extraction rather than prediction, we add another model $h_\phi$ for classification. The training objective then becomes:
\begin{equation}\label{eq:objective_for_stage1}
    \min_\theta \sum_{n=1}^{2N} \mathcal{L}(h_\phi(f_\theta(t_{i_{n_1}}, t_{i_{n_2}})), y_n) 
\end{equation}
where $(t_{i_{n_1}}, t_{i_{n_2}}))$ is the corresponding feature pair of $(i_{n_1}, i_{n_2}), n\in \{1,\cdots, 2N\}$. And $y_n$ denotes whether the $n$-th pair is positive ($y_n=1$) or negative ($y_n=0$). 
Once we jointly train $f_\theta$ and $h_\phi$ using the dataset, we obtain a feature extractor that can identify significant features, and $h_\phi$ can determine whether the pair is a good match. If the pair is deemed a good match, it will proceed to the next stage, where explanations will be generated.

\begin{figure*}
    \centering
    \includegraphics[width=\linewidth]{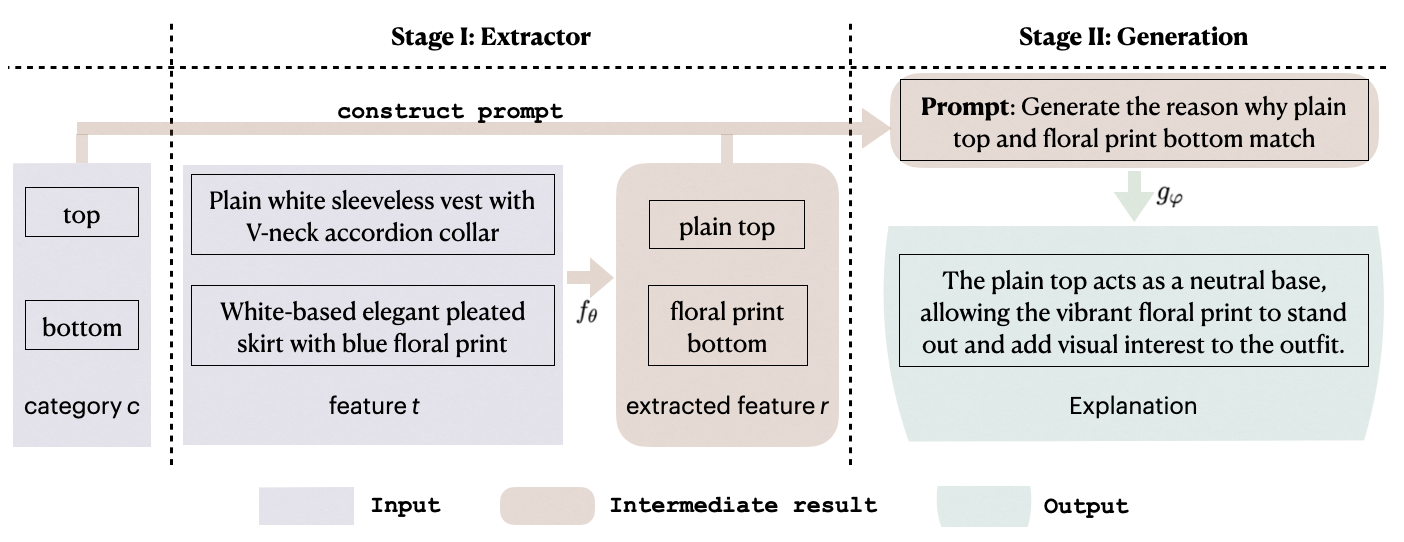}
    \caption{Overall Proposed Pipeline. The features $t$ (in the form of bag of words) of two items are input into the extraction model $f_\theta$ (without the category) to obtain the extracted features $r$. Then we construct the prompts with the category $c$ and the extracted features $r$, which are input into the generation model $g_\varphi$, yielding the final explanation.}
    \label{fig:pipeline}
\end{figure*}

\subsection{Stage II: Generate Explanations}
\label{ssub: stage2_generate_explanations}
Next
we want to generate reasonable explanations for the item pair $(i, j)$ with the categories $(c_i, c_j)$ and the extracted features $(r_i, r_j)$ in Section \ref{ssub: stage1_extract_important_features}. 
However, supervision in this stage is also missing. So we first construct \textbf{PFE}-dataset, where \textbf{PFE} stands for Pair Fashion Explanation, which will be used for finetuning our model.

\subsubsection{PFE-dataset Construction}
We manually define a set of key paraphrases: \{"Clothes Match", "Clothes Fashion", "Fashion", 
"Outfit of The Day", "Style", "Match Clothes", "How to Dress", "How to Wear", "Match"\}. For each key paraphrase, we conduct a search on public magazines and crawl all related articles. 
This yields a dataset of 959,157 sentences. However, not all of these sentences are relevant to our task of pair matching. 
\wy{To filter out irrelevant sentences, we adopt the package \texttt{spacy} to perform named entity recognition (NER) on the sentence, extracting the entities with the NER tag ``Noun".}
We create a dictionary of relevant entities (shown in Appendix \ref{dictionary_of_categories}) and extract sentences containing more than two such entities, resulting in a reduced dataset of 48,726 sentences. We then turned to ChatGPT to extract important features from the sentences. 
With the prompts and detailed process shown in Appendix \ref{prompts_to_query_chatgpt}, the sentences that do not necessarily explain the relationship between items are filtered out.
We end up with a final dataset of 6,407 sentences, each with important features, categories, and ground truth explanations denoted by $\{r_i, r_j, c_i, c_j, t_e\}$. For instance, one example in this dataset is: 
\vspace{3pt}\\
$r_i, r_j$: \texttt{skirt, belt};\\
$c_i, c_j$: \texttt{kilt, studded};
\vspace{2pt} \\
$t_e$\texttt{\footnotesize \mbox{:The outfit looks cohesive because} the oversized layers are cinched with a studded belt, which complements the little strip from a kilt skirt that is also affixed to the belt, creating a visually pleasing balance in the outfit.}

\wy{\noindent We also report the percentages of each item in this dataset. For every item, we calculate the percentage of each item appearing in the whole dataset, shown in Table \ref{tab:dictionary_of_categories_with_percentage}}. 

\subsubsection{Explanations Generation}
In the domain of natural language processing, there is the general routine of training the model on a large corpus and then finetuning it on the domain-specific dataset~\citep{SPDF,MatSciBERT}.
Thus we utilize a pre-trained language model to fine-tune it on our PFE-dataset. The idea is that we could create a template to construct the prompt as input to the language model, with the target sentences being the output. 

\noindent Specifically, the objective of training is:
\begin{equation}
    \max P_{g_\varphi}(t_e|\textrm{template}(r_i, r_j, c_i, c_j))
\end{equation}
We denote $x_s$ as the generated word at step $s$, then the objective becomes:
\begin{equation}
    \max \sum_{s=1}^{|t_e|}P_{g_\varphi}(x_s|\textrm{template}(r_i, r_j, c_i, c_j) + x_{:s})
\end{equation}
After training on the PFE-dataset, we input the features extracted from Stage I into the model $g_\varphi$ to yield the explanations for the features in Stage I. The overall framework of our method is shown in Figure \ref{fig:pipeline}.

\wy{
\subsection{Instantiation of Models}
Our framework is model-agnostic, suitable for various extraction tasks in Stage I, and adaptable to different large language models in Stage II. This flexibility allows for diverse model implementations, as demonstrated by our selected examples:  

\subsubsection{Models for Stage I}
Various extraction models could serve as $f_\theta$; we discuss two specific implementations:

\paragraph{Cross-Attn}
We estimate the attention scores for each pair given $t_i$ and $t_j$, then compute the weighted average of all concatenated pair embeddings. The resulting vector inputs into an MLP to produce the prediction. Implementation details are available in Appendix \ref{cross_attn_implementation}.

\paragraph{Rationale Extraction} 
Provided $t_i$ and $t_j$ and their respective binary labels, we utilize a rationale extraction approach to highlight the most significant words from each sentence. Specifically, we concatenate $t_i$ and $t_j$ into a single string $t$ separated by the \texttt{<end>} token. Following the rationale extraction paradigm~\cite{RE,Kuma,DebiasRE}, we train an extractor $f_\theta$ and a classifier $h_\phi$. The trained $f_\theta$ extractor serves as the extraction model in our pipeline (Figure \ref{fig:pipeline}).

\subsubsection{Models for Stage II}
\label{models_for_stage_2}
Following the pretraining-finetuning routine in the state-of-the-art methods of explanation generation in recommendation~\citep{PEPLER,li23unifying}, 
we conduct experiments with the representative language models listed below:

\paragraph{GPT-2}
GPT-2, a large transformer-based language model with 124 million parameters~\cite{GPT2}, can generate simple, albeit not necessarily comprehensive, sentences. GPT-2 serves as a basic baseline in our study.

\paragraph{Flan-T5-large}
Flan-T5~\cite{flan-t5} is an augmented version of T5~\cite{T5} and is trained on chain-of-thought data. The Flan-T5-large model is one of several models in the Flan-T5 series, boasting 780 million parameters.

\paragraph{Flan-T5-xl}
Flan-T5-xl, also part of the Flan-T5 series, comprises 
3 billion parameters. 

}





%% file: 4_experiments.tex
\section{Experiments}
\subsection{Experimental Setup}

\subsubsection{Datasets for Stage I}
To train models $f_\theta$ and $h_\phi$ in accordance with Eq.\eqref{eq:objective_for_stage1}, we construct a composite dataset \textbf{Combined} from \textbf{Amazon Reviews}\citep{ni2019justifying} and \textbf{FashionVC}\citep{neurostylist}.

The Amazon Review dataset comprises numerous subsets; we focus on the largest subset related to fashion, specifically ``Clothing, Shoes and Jewelry". For each item, we leverage the "title" attribute and the "also\_buy" list to build pairs $(i, j)$, where $i$ and $j$ are individual items. We extract the categories $c_i$ and $c_j$ via manual noun dictionary matching, treating the remaining text as the features $t_i$ and $t_j$ used in Eq. (\ref{eq:extraction_model}). In this way, we force the model to determine if it is a good pair based on the features rather than the categories. By excluding pairs where both items belong to the same category, our dataset ultimately contains 124,679 pairs with 77,113 items.

The FashionVC dataset includes 20,715 pairs of tops and bottoms, consisting of 14,871 tops and 13,663 bottoms. Here, each top or bottom possesses a corresponding category and feature, which we utilize as $c_i$ and $t_i$ for training.

\paragraph{Combined}
Recognizing the similar knowledge embedded within the two datasets, we integrate them into a single dataset, yielding a total of 145,394 pairs and 105,647 unique items. To generate negative samples, for each positive pair $(i,j)$, we randomly select another item $j'$, provided that $(i,j')$ is absent from the positive pair set and $i$ and $j'$ belong to distinct categories. The final dataset 
includes
145,394 negative pairs.


\subsubsection{Evaluation Metrics}
For automated assessment, following \citet{PEPLER},  we 
use
BLEU scores~\citep{BLEU} and Rouge scores~\citep{rouge} to evaluate the quality of the generated text, BLEURT~\citep{BLEURT} to compare the semantic similarity of our generated sentences to human-written sentences. Additionally, we use the Frechet Inception Distance (FID)~\citep{fid} to measure the consistency between the distribution of our model's output and the original sample distribution. We use CLIP~\citep{clip} ViT-B/32 text encoder to encode the texts before calculating the distribution distance. 

\subsection{Implementation Details}
\subsubsection{Stage I Details}
We partition the Combined dataset into training, validation, and test sets at a ratio of 8:1:1. We jointly train the extractor $f_\theta$ and the classifier $h_\phi$ as depicted in Eq.\eqref{eq:objective_for_stage1} to acquire the extractor $f_\theta$. For Cross-Attn, we apply lasso regularization with a weight value of 0.01. For Rationale Extraction, we fix the selection ratio at 0.3, indicating we select 30\% of text from $t_i$ and $t_j$ as the rationale.

\subsubsection{Stage II Details}
We employ prompts ``$f_i$ $c_i$ and $f_j$ $c_j$ match because" and ``Generate the reason why $f_i$ $c_i$ and $f_j$ $c_j$ match:" for GPT2 and Flan-T5, respectively. For GPT2, we set the batch size as 5 to train the whole model. While for Flan-T5-large and Flan-T5-xl, we use the package \texttt{peft}~\citep{peft} with the method LoRA~\citep{lora} to finetune. 
For Flan-T5-large, we set the batch size as 5. For Flan-T5-xl, we set the batch size as 1 but accumulate the gradient over 5 iterations. The learning rate for all models is set to 2e-4. The whole dataset is split into training set and testing set with the ratio of 9:1. 
The held-out testing set is also used for the evaluation in Table \ref{tab:overall_performance_comparison}.

\subsection{Overall Performance Comparison}
We present a comprehensive comparison of our model's performance against several baseline models, which fall into the following groups:

\vspace{2pt}
\noindent \textbf{Naive Repetition}: This model merely repeats the input features, hence offering a simple benchmark.

\vspace{2pt}
\noindent \textbf{Recommendation Models}: Models such as PETER~\cite{Peter} and PEPLER~\cite{PEPLER}, originally designed for explaining item recommendations, are repurposed for this task with slight modifications and finetuning on the PFE dataset.

\vspace{2pt}
\noindent \textbf{Large Language Models}: General-purpose language models like GPT2, Flan-T5-large, Flan-T5-xl, and ChatGPT, are employed without any specific fine-tuning on the PFE-dataset. For ChatGPT, we use the version \texttt{gpt-3.5-turbo} for comparison.

\vspace{2pt}
\noindent \textbf{Ours}: GPT2, Flan-T5-large, and Flan-T5-xl finetuned on the PFE-dataset training set.

\begin{table*}[t]
    \centering
    \resizebox{0.92\linewidth}{!}{%
    \begin{tabular}{cccccccc}
    \toprule
        Model & BLEURT & BLEU-1 & BLEU-4 & ROUGE-1 & ROUGE-2 & ROUGE-L & FID \\
    \midrule 
        Naive Rep & -0.3731 & 27.35 & 5.78 & 0.3963 & 0.1391 & 0.2849 & 8.670 \\
    \midrule
    PETER & -1.2367 & 0.02 & 0.00 & 0.0005 & 0.0000 & 0.0005 & 60.004 \\
    PEPLER-F & -0.3115 & 31.52 & 3.76 & 0.2892 & 0.0586 & 0.2327 & 10.313 \\
    \midrule
        GPT-2 & -0.6656 & 10.47 & 0.17 & 0.1039 & 0.0051 & 0.0822 & 33.847 \\ 
        Flan-T5-large & -0.6398 & 15.33 & 3.16 & 0.3194 &0.1154 &  0.2349 & 9.047 \\
        Flan-T5-xl & -0.5783 & 23.64 & 4.37 & 0.2825 & 0.1004 & 0.2179 & 9.634 \\
        ChatGPT & 0.0139 & 35.67 & 6.34 & 0.3872 & 0.1274 & 0.2739 & 5.802 \\
    \midrule
        GPT-2-F & -0.0853 & 38.18 & 9.01 & 0.3868 & 0.1442 & 0.2990 & 3.014 \\
        Flan-T5-large-F & 0.0413 & 43.80 & 12.57 & 0.4557 & 0.2081 & \textbf{0.3538 }& \textbf{2.256} \\
        Flan-T5-xl-F & \textbf{0.0736} & \textbf{46.12} & \textbf{14.11} & \textbf{0.4691} & \textbf{0.2208} & 0.3485 & 3.046  \\
    \bottomrule
    \end{tabular}}
    \caption{Overall Performance Comparison on the held-out test set of PFE dataset, measuring the alignment between the generated explanations and the ground-truth compatibility relationships. ``Naive Rep'' means "Naive repetition". ``F'' means being fine-tuned with the PFE dataset. }
    \label{tab:overall_performance_comparison}
\end{table*}

\subsubsection{Automatic Evaluation}
\paragraph{Evaluation with PFE-dataset}
Our evaluation 
is predominantly based on
the PFE dataset test set. For each data instance, the model generates an explanation whose distance from the ground truth explanation is measured and summarized. The results presented in Table \ref{tab:overall_performance_comparison} illustrate several key findings: (1) Our model, when finetuned on the PFE-dataset, exhibits superior performance in generating explanations that align closely with the ground truth. This underlines the value of training language models specifically for explanation generation tasks. (2) Despite recommendation models being domain-specific, their adaptations struggle with multi-object explanation generation. This underscores the complexities and unique aspects of explanation generation that necessitate a distinct, tailored approach. (3) Large language models, although capable of generating semantically coherent sentences and capturing entities from prompts (evident from high BLEU-1 scores), fall short in generating highly relevant explanations, as evidenced by their relatively lower BLEURT scores. This discrepancy highlights the pivotal role of training data like the PFE-dataset, which provides specific ground-truth compatibility explanations, thereby enhancing model performance in generating more accurate and meaningful explanations.

\paragraph{Evaluation for whole process}

Moreover, we evaluate our two-stage framework by comparing it with other models that lack an extraction stage. Here, the evaluation metric is the FID between generated explanations and the PFE-dataset. Table \ref{tab:fid_for_whole_process} demonstrates that our models yield significantly lower FID scores, signifying that our models are more adept at producing explanations that resemble the distribution of the ground truth dataset. The results suggest that our models are generating explanations, not merely describing the two items, which is crucial for the task at hand. 

\subsubsection{Human Evaluation}

Our human-centric evaluation involves comparing the entire framework with the top-performing models identified earlier. From the Combined dataset, we randomly selected 50 pairs and generated explanations using PEPLER, ChatGPT, and our fine-tuned Flan-T5-xl with the Rationale Extraction model. 
We posed the task as a rating problem for explanations, where each explanation was rated on a scale of 1-10. Three Amazon Turk workers, with a minimum \texttt{HIT Approval Rate} of 80\%, independently rated each explanation. The average rating for each model is reported in Table \ref{tab:overall_human_evaluation}. 
Our model consistently outperforms PEPLER and maintains competitive performance with ChatGPT, which is widely recognized for its capacity to generate fluent and coherent sentences. The fact that our model can match the performance of ChatGPT suggests that it is not only producing linguistically competent explanations but is also effectively capturing and conveying the complex relationship between the pair of items.

\begin{table*}[ht]
    \centering
    \resizebox{\linewidth}{!}{%
    \begin{tabular}{c|c|c|cccc|ccc}
        \toprule
          & Naive Rep & PEPLER & GPT2-RE & Flan-T5-large-RE & Flan-T5-xl-RE & ChatGPT & GPT2-F-RE & Flan-T5-large-F-RE & Flan-T5-xl-F-RE \\
         \midrule
        FID & 20.657 & 14.918 & 40.423 & 20.086 & 23.951 & 18.667 & 7.689 & 8.091 & 11.657 \\
        \bottomrule
    \end{tabular}}
    \vspace{-4pt}
    \caption{Comparison of FID for various models. ``F'' refers to being fine-tuned, ``RE'' means the model is equipped with the Rationale Extraction (RE) component as the extractor. }
    \label{tab:fid_for_whole_process}
\end{table*}

\begin{table}[]
    \centering
    \resizebox{\linewidth}{!}{%
    \begin{tabular}{c|ccc}
        \toprule
        Model & PEPLER & ChatGPT & RE + Flan-T5-xl-F \\
        \midrule
        Avg Rating & 5.48 & \textbf{6.50} & 6.43 \\
        \bottomrule
    \end{tabular}}
    \caption{Human Evaluation. }
    \label{tab:overall_human_evaluation}
\end{table}

\subsection{Ablation Study}
\subsubsection{Ablation Study for Stage I}
\paragraph{Overall Performance of the Extraction Model} 
The 
performance
of the Cross-Attn and Rationale Extraction models are detailed in Table \ref{tab:accuracy_stage_1}. Both models display an ability to discern compatibility correlations, with the Rationale Extraction model performing slightly better. This could be due to its more complex LSTM structure.

\begin{table}[t]
    \centering
    \resizebox{\linewidth}{!}{%
    \begin{tabular}{c|c|cccc}
        \toprule
        Method & Acc & R@5 & R@10 & R@20 & R@50 \\
        \midrule
        Cross-Attn &  0.8505 & 0.1468 & 0.2229 & 0.3187 & 0.4846 \\
        Rationale Extraction & 0.8531 & 0.3060 & 0.3984 & 0.5049 & 0.6596  \\ 
        \bottomrule
    \end{tabular}}
    \vspace{-4pt}
    \caption{Performance of different models in Stage I.}
    \vspace{-10pt}
    \label{tab:accuracy_stage_1}
\end{table}

\paragraph{Influence of Extraction Models on Explanation Generation Fidelity}
We investigated the effect of the extraction models on the overall quality of generated explanations by calculating the FID between the CLIP~\cite{clip} embeddings of the generated explanations and the ground-truth explanations. We adopt CLIP ViT-L/14 for the calculation. The results are presented in Table \ref{tab:FID_With_different_extraction_models}. Two key observations can be made from the results: (1) All models equipped with extraction methods were capable of producing explanations more closely resembling ground-truth explanations. (2) Interestingly, the finetuned GPT2 model generated explanations that have the smallest FID. Although this only suggests that GPT2 can better mimic the pattern of the training corpus rather than identifying compatibility relationships, it nevertheless implies potential utility for future studies.


\begin{table}[ht]
    \centering
    \resizebox{\linewidth}{!}{%
    \begin{tabular}{cccc}
    \toprule
          & None & Cross-Attn & Rationale Extraction \\
    \midrule
       GPT2-F & 7.980 & \textbf{6.718} & 7.689 \\
       Flan-T5-large-F & 11.141  & \textbf{8.073} & 8.091 \\
       Flan-T5-xl-F & 16.647 & \textbf{10.518} & 11.657  \\
    \bottomrule
    \end{tabular}}
    \caption{FID with different extraction models.}
    \label{tab:FID_With_different_extraction_models}
    \vspace{-12pt}
\end{table}

\begin{table}[ht]
    \centering
    \resizebox{\linewidth}{!}{%
    \begin{tabular}{c|cccc}
        \toprule
        Model & GPT2 & Flan-T5-large & Flan-T5-xl & ChatGPT \\
        \midrule
        Avg Rating & 5.91 & 6.27 & \textbf{6.51} & 6.37  \\
        \bottomrule
    \end{tabular}}
    \vspace{-4pt}
    \caption{User study for different generation models.}
    \label{tab:user_study_for_different_generation_models}
    \vspace{-5pt}
    \vspace{-12pt}
\end{table}

\subsubsection{Ablation Study for Stage II}
\label{ssub:ablation_study_for_stage2}

\begin{table*}[t!]
    \centering
    \begin{tabular}{c|p{14.5cm}}
\toprule
\textbf{Item/Feature} & \textbf{boot/timberland men groveton ltt chukka; shirt/ularmo men boy summer cotton tees short} \\
\midrule
\textbf{PEPLER-F} & the black color of the shirt complements the black color of the pants, creating a cohesive color scheme. the black color of the shirt also ties the whole outfit together.\\
\midrule
\textbf{ChatGPT}  & The Timberland men's Groveton LTT Chukka Boot and Ularmo Men Boy Summer Cotton Tees Short Star Printed T L Black Shirt Sleeve Shirt match because the casual yet stylish design of the boots complements the graphic print of the t-shirt. \\
\midrule
\textbf{Flan-T5-xl-F-RE}  & The Timberland ltt chukka boot and ularmo summer cotton tees shirt sleeve shirt complement each other because they both have a casual and laid-back vibe, creating an effortless outfit.\\
\bottomrule
    \end{tabular}
    \caption{Our model Flan-T5-xl is equipped with the Rational Extraction model from the first stage, and ``F'' means fine-tuned with PFE-dataset. ChatGPT and PEPLER-F take the raw features shown in the first row as input.}
    \label{tab:cases_overall}
\end{table*}

\begin{table*}[t!]
    \centering
    \begin{tabular}{c|p{14.5cm}}
\toprule
\textbf{Item/Feature} & 
\textbf{Jacket and mini skirt/Matching black tweed cropped; heels/vibrant pink with oversized black bows on the back} \\
\midrule
\textbf{GPT2-F} &  The white tee with oversized black bows on the back match the black tweed cropped jacket, and the vibrant pink heels add pops of contrasting color to tie the whole outfit together. \\
\midrule
\textbf{Flan-T5-large-F} & The black tweed jacket and mini skirt match each other in terms of color and texture, while the vibrant pink heels with oversized black bows on the back add a pop of color that complements the black and pink tones in the outfit. Overall, the combination creates a stylish look.\\
\midrule
\textbf{Flan-T5-xl-F} & The black tweed cropped jacket and mini skirt create a cohesive look, while the vibrant pink heels with oversized black bows on the back add a pop of color to the outfit.\\
\midrule
\textbf{ChatGPT} & The reason why a matching black tweed cropped jacket and miniskirt pair well with vibrant pink heels featuring oversized black bows on the back is that the combination creates a bold and stylish contrast, combining classic elegance with playful accents. \\
\midrule
\textbf{Ground Truth} & The black tweed fabric of the jacket and mini skirt creates a cohesive outfit, while the vibrant pink slingback heels with the oversized black bows add a pop of color and playful touch to the overall look. \\
\bottomrule
    \end{tabular}
    \caption{Generation results in testset of PFE dataset of Stage II. ``F'' means fine-tuned with PFE-dataset.}
    \label{tab:cases_in_stage2}
\end{table*}

\paragraph{Human Evaluation of Generation Model with Stage I Keywords}
Upon the successful training of Stage I and Stage II, we conduct the generation with the models $f_\theta$ and $g_\phi$ by inputting the extracted features (with $f_\theta$) of true positive pairs into the Stage II generation model ($g_\phi$). The rationale extraction model served as our primary extractor.

We also engaged ChatGPT in our experiment. For each item pair characterized by features $f_i$, $f_j$ and categories $c_i, c_j$, we queried ChatGPT with the following question:\\
\vspace{-7pt}

\texttt{What is the reason why $f_i$ $c_i$ and $f_j$ $c_j$ match, and could you please provide a concise response (with one or two sentences) that can be directly shown to customers?}\\
\vspace{-7pt}

Post explanation generation from different models, we carried out a user study as detailed below:

\paragraph{Evaluation of Generation}: We randomly selected 50 prompts from the first stage and leveraged GPT2, Flan-T5-large, Flan-T5-xl (all finetuned), and ChatGPT to generate explanations. Subsequently, for each item pair, we asked the users to score the explanations on a 1-10 scale, focusing on their conciseness, persuasiveness, and overall acceptability, which would convince them that the provided pair of items indeed form an excellent match. We delegated each question to three Amazon Turk workers, maintaining a 
criteria that their \verb|HIT Approval Rate| must exceed 80\%. Table \ref{tab:user_study_for_different_generation_models} summarizes the results. From this table, two observations can be made: (1) Flan-T5-xl outperformed ChatGPT when the extracted features from Stage I were used, showcasing our model's ability to generate more user-acceptable explanations. (2) Flan-T5-xl, compared with Flan-T5-large and GPT2, recorded the best performance, indicating that larger models possess superior abilities to internalize knowledge from the finetuned dataset.

\begin{figure}[ht]
\centering
\subfigure[BLEURT]{\label{fig:bleu}\includegraphics[width=0.480\linewidth]{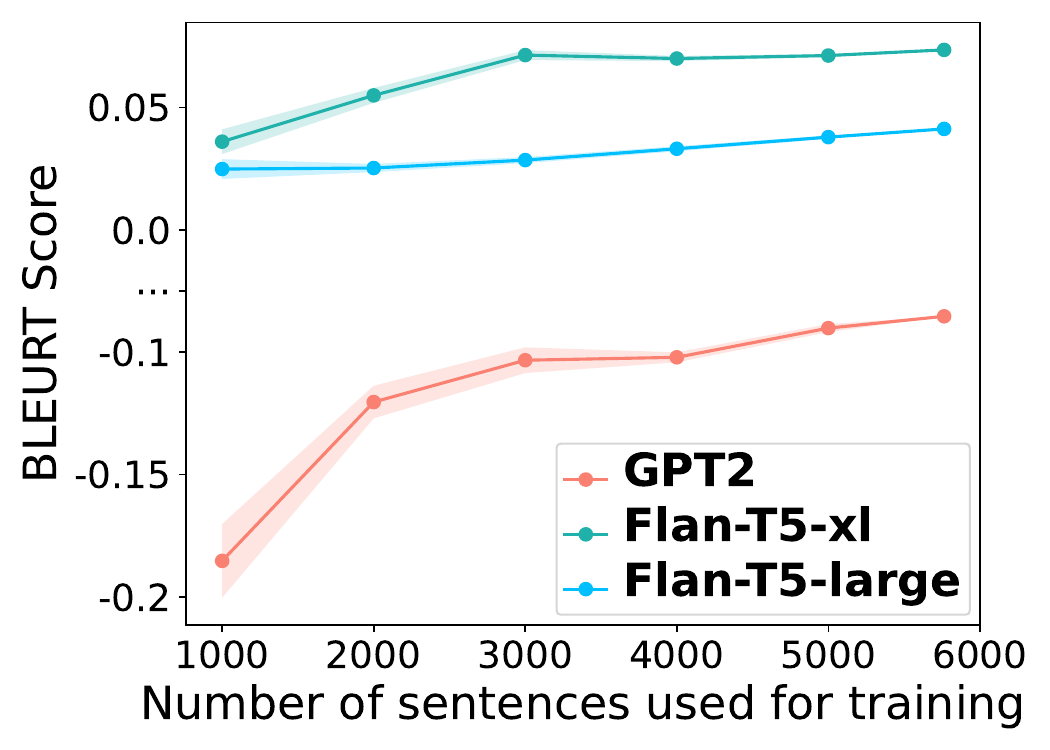}}
\hfill
\subfigure[Rouge-2]{\label{fig:rouge2}\includegraphics[width=0.480\linewidth]{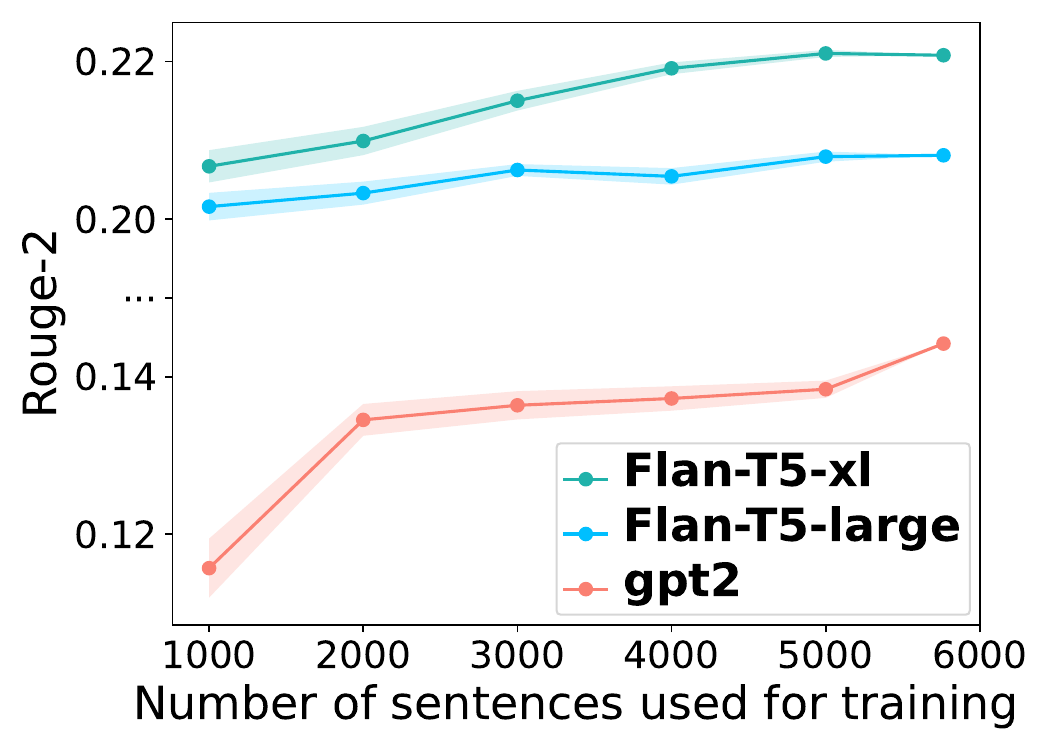}}
\caption{Effect of the number of sentences.}
\label{fig:effect_of_number_of_sentences}
\vspace{-10pt}
\end{figure}

\paragraph{Influence of Training Sample Size}
To investigate whether the PFE-dataset, consisting of 6,405 examples, is sufficient for the model to discern the compatibility relationships between the paired items, we performed experiments by finetuning the LLM with a subset of the PFE-dataset. The results are illustrated in Figure \ref{fig:effect_of_number_of_sentences}. From this figure, it can be concluded that: (1) Larger models necessitate fewer samples for finetuning. The smallest model, GPT2, exhibits enhanced performance with an increasing number of samples, while the gains for Flan-T5-large are relatively modest when the sample size increases. As for Flan-T5-xl, the BLEURT and Rouge-2 scores barely rise when samples are increased from 5,000 to 6,000. (2) Although 6,405 samples (5,764 samples used for training) might appear to be limited, the results suggest that further sample augmentation may not significantly boost the performance. 


\subsection{Case Study}

\subsubsection{Illustrative Examples after Combining Stage I and Stage II}
We show generated explanations from PEPLER, ChatGPT, and our Flan-T5-xl model in Table \ref{tab:cases_overall}. In the example, we find that the recommendation model PEPLER may fall short in terms of the fluency of the explanations. Both ChatGPT and Flan-T5-xl capture the concept of ``casual'', showing that our model can compete favorably with ChatGPT. This aligns with the quantitative results in Table \ref{tab:overall_human_evaluation} where the rating of our model is similar to the rating of ChatGPT.

\subsubsection{Illustrative Examples from Stage II}
In Stage II, we present generated examples in Table \ref{tab:cases_in_stage2}. These are selected from the test set of the PFE-dataset, with additional results available in the Appendix. From the table, we could observe that our models Flan-T5-large-F and Flan-T5-xl-F mention the crucial point that the pink heels add "a pop of color", a detail that ChatGPT fails to touch upon. This confirms that although ChatGPT may excel at sentence construction, it might miss capturing important compatibility relations between items. This shows that our model might be more sensitive to the domain-specific relations compared with general-purpose LLMs.

%% file: 5_conclusion.tex
\section{Conclusion}

In this paper, we addressed the challenge of generating explanations for the compatibility of paired fashion items, an area that has seen limited research focus to date. The value of our work is twofold. First, we have constructed the PFE-dataset that can be leveraged for further research in Fashion Recommendation. Secondly, our framework offers a generalized approach for revealing compatibility relationships, extending its potential application beyond the confines of our initial focus on fashion items. Notably, our framework is not limited to the realm of fashion; it also has potential applications in determining the compatibility of various pieces of furniture for interior design purposes and establishing the interoperability of different electronic devices. 
A current limitation of our method is that it may find multi-object compatibility relationships challenging to reveal. Thus in the future, we aim to extend our framework into explaining multi-object compatibility relationships.
The applications of our approach include curating collections of fashion items or coordinating comprehensive interior designs. 
In conclusion, our work serves as a stepping stone towards a better understanding of the underlying factors that contribute to the compatibility of items, and offers a versatile tool that holds promise for wider application and further refinement.

%% file: 7_appendix.tex
\section{Filtering Process for PFE-dataset}

\begin{table*}[]
    \centering
    \begin{tabular}{p{15cm}}
\noindent \texttt{
 \{'dress', 'top', 'skirt', 'jacket', 'shirt', 'pant', 'boot', 'jean',
       'jeans', 'bag', 'coat', 'suit', 'trouser', 'blazer', 'trousers',
       'sweater', 'shoe', 'blouse', 'shorts', 'shoes', 'sneaker',
       'sandal', 'belt', 'pump', 'hat', 'scarf', 'leggings', 'necklace',
       'bra', 'vest', 'cardigan', 'gown', 'loafer', 'sock', 'sunglass',
       'handbag', 'sweatshirt', 'bodysuit', 'miniskirt', 'velvet', 'tote',
       'satin', 'wool', 'jumpsuit', 'cloth', 'bracelet', 'plaid', 'cap',
       'hoodie', 'corset', 'block', 'uniform', 'watch', 'cover', 'jumper',
       'sundress', 'robe', 'clothes', 'pack', 'bustier', 'swimsuit',
       'bootie', 'overalls', 'clog', 'shawl', 'slipper', 'lingerie',
       'beret', 'fedora', 'pullover', 'costume', 'slingback',
       'sweatpants', 'beanie', 'backpack', 'lounge', 'ballerina',
       'espadrille', 'panty', 'windbreaker', 'kilt', 'waistcoat',
       'leotard', 'saddle', 'brogue', 'pantyhose', 'jumpsuits', 'culotte',
       'pouch', 'kimono', 'caftan', 'moccasin', 'bloomer', 't-shirt',
       'briefcase', 'visor', 'sari', 'underwear', 'wallet', 'cloche',
       'duffel', 'swimwear', 'panama', 'slip-on', 'ballgown', 'satchel'\}
}
    \end{tabular}
    \caption{Manually curated dictionary containing all categories.}
    \label{tab:dictionary_of_categories}
\end{table*}

\begin{table*}[]
    \centering
    \begin{tabular}{p{15cm}}
\noindent \texttt{
 \{dress: 9.10\%; top: 6.99\%; skirt: 6.89\%; jacket: 6.01\%; shirt: 4.99\%; pant: 4.64\%; boot: 4.12\%; jean: 4.01\%; jeans: 3.79\%; bag: 3.11\%; coat: 3.07\%; suit: 2.94\%; trouser: 2.77\%; blazer: 2.75\%; trousers: 2.50\%; sweater: 2.45\%; shoe: 2.20\%; blouse: 1.97\%; shorts: 1.86\%; shoes: 1.77\%; sneaker: 1.70\%; sandal: 1.58\%; belt: 1.50\%; pump: 1.39\%; hat: 0.98\%; scarf: 0.91\%; leggings: 0.84\%; necklace: 0.81\%; bra: 0.79\%; vest: 0.77\%; cardigan: 0.77\%; gown: 0.67\%; loafer: 0.65\%; sock: 0.63\%; sunglass: 0.60\%; handbag: 0.45\%; sweatshirt: 0.41\%; bodysuit: 0.40\%; miniskirt: 0.38\%; velvet: 0.34\%; tote: 0.27\%; satin: 0.27\%; wool: 0.27\%; jumpsuit: 0.26\%; cloth: 0.25\%; bracelet: 0.23\%; plaid: 0.22\%; cap: 0.21\%; hoodie: 0.21\%; corset: 0.21\%; block: 0.17\%; uniform: 0.17\%; watch: 0.17\%; cover: 0.14\%; jumper: 0.14\%; sundress: 0.13\%; robe: 0.13\%; clothes: 0.13\%; pack: 0.11\%; bustier: 0.11\%; swimsuit: 0.11\%; bootie: 0.10\%; overalls: 0.09\%; clog: 0.09\%; shawl: 0.08\%; slipper: 0.08\%; lingerie: 0.08\%; beret: 0.08\%; fedora: 0.08\%; pullover: 0.07\%; costume: 0.06\%; slingback: 0.06\%; sweatpants: 0.05\%; beanie: 0.05\%; backpack: 0.05\%; lounge: 0.04\%; ballerina: 0.04\%; espadrille: 0.04\%; panty: 0.04\%; windbreaker: 0.04\%; kilt: 0.03\%; waistcoat: 0.03\%; leotard: 0.03\%; saddle: 0.02\%; brogue: 0.02\%; pantyhose: 0.02\%; jumpsuits: 0.02\%; culotte: 0.02\%; pouch: 0.02\%; kimono: 0.02\%; caftan: 0.01\%; moccasin: 0.01\%; bloomer: 0.01\%; t-shirt: 0.01\%; briefcase: 0.01\%; visor: 0.01\%; sari: 0.01\%; underwear: 0.01\%; wallet: 0.01\%; cloche: 0.01\%; duffel: 0.01\%; swimwear: 0.01\%; panama: 0.01\%; slip-on: 0.01\%; ballgown: 0.01\%; satchel: 0.01\% \}
}
    \end{tabular}
    \caption{Manually curated dictionary containing all categories and the corresponding percentages. }
    \label{tab:dictionary_of_categories_with_percentage}
\end{table*}

\subsection{Dictionary of Categories}
\label{dictionary_of_categories}
The dictionary extracted from Amazon categories and revised manually is shown in Table \ref{tab:dictionary_of_categories}. The percentages of these items are shown in Table xxx.

\subsection{Prompts to query ChatGPT}
\label{prompts_to_query_chatgpt}
After filtering with \verb|spacy| package, we obtain 48,726 sentences. Then we turn to GPT-3.5-turbo for filtering. The prompt for the first cycle of filtering is as follows: \\

\noindent \texttt{
\noindent Let me give you an example:\\
Description: The hits run from Poirets stunning 1919 opera coat, made of a single swath of uncut purple silk velvet, to a 2018 kimono printed with oversize manga characters by Comme des Garons founder Rei Kawakubo.\\
Items: coat, kimono;\\
Key features:\\
- coat: purple silk velvet;  \\
- kimono: oversize manga characters; \\
Then please give the key features concisely following the above structure in the following question: \\
Description: \{\textrm{sentence}\} \\ 
Items: \{\textrm{items}\} \\ 
Key features:\\ 
}

where the sentence and items are the sentence from the dataset and the entities extracted with \verb|spacy|. 

After querying ChatGPT, we end up with all the features corresponding to the extracted items, where some of key features are denoted as ``not-specified", meaning that ChatGPT could not find the key features. Such sentences are dropped. Then we have 37737 sentences left. 

Then we query ChatGPT again with the following prompt:\\

\texttt{
Could you read this sentence and let me know if it is explaining why two pieces of clothing look good together as an outfit? It's possible that the sentence could be one or several sentences about why the two items complement each other and create a cohesive outfit. If no, then simply answer a "No"; If yes, please give a concise reason for how they complement each other in the form of "Reason: They match because ...".\\
\{sentence\}\\
}

where the sentence is the extracted sentence from the above 37737 sentences. With the returned answers, we filter out the sentence with the answer "No". For the left sentences, we construct a new dataset with the entities, and features extracted from the previous and the sentences rewritten by ChatGPT with the answer "Yes". After the above process, we obtain a dataset with 6,407 examples.

\section{Implementation Details}

The configurations for fine-tuning adapted languages models are: learning rate=0.0002, weight decay=0, optimizer=Adam, training epoch=20, batch size=5 for GPT2 and Flan-T5-large, while batch size=1 for Flan-t5-xl, max length=100, 
All codes are implemented with Python3.9.12 and PyTorch2.0.1 with CUDA 11.7. operated on Ubuntu (16.04.7 LTS) server with 2 NVIDIA GeForce GTS A6000 GPUs. Each has a memory of 49GB.

\subsection{Cross Attention Extractor}
\label{cross_attn_implementation}
With $t_i$ and $t_j$, we use \verb|Bert-tokenizer|~\cite{bert} to tokenize the sentences and get the embedding for each word $\{\textbf{e}_{i1},\cdots,\textbf{e}_{il_i}\}$ and $\{\textbf{e}_{j1},\cdots,\textbf{e}_{jl_j}\}$, where $l_i$ and $l_j$ are the lengths of the indices of tokenized sentences $t_i$ and $t_j$. $\{\textbf{e}_{i1},\cdots, \textbf{e}_{il_i}\}$, $\{\textbf{e}_{j1},\cdots,\textbf{e}_{jl_j}\}$ are the embeddings for $t_i$ and $t_j$, respectively, with each $\textbf{e} \in \mathbb{R}^{512}$. 
Then we calculate the attention map $\mathbf{A}\in\mathbb{R}^{l_i\times l_j}$ with each element as:
\begin{align}
    a_{k_ik_j} &= \textrm{Sigmoid}(\textrm{MLP}(\textrm{Cat}(\textbf{e}_{ik_i}, \textbf{e}_{jk_j}))), \label{eq:ftheta_for_cross_attn} \\
    k_i&\in\{1,\cdots,l_i\}, k_j \in \{1,\cdots,l_j\} \nonumber
\end{align}
Then we normalize the attention score so that all scores add up to 1: 
\begin{equation}
    \overline{\mathbf{A}} = \mathbf{A} / \textrm{Sum}(\mathbf{A})
\end{equation}
where $\textrm{Sum}(\mathbf{A})$ means the sum of all the elements in $\mathbf{A}$. Then we could get the weighted average of the concatenated embeddings:
\begin{equation}
    \textbf{e}_{\mathit{avg}} = \frac{1}{l_i * l_j} \sum_{k_i=1}^{l_i} \sum_{k_j=1}^{l_j} \overline{a}_{k_ik_j} \textrm{Cat}(\textbf{e}_{k_i}, \textbf{e}_{k_j})
\end{equation}
where $\overline{a}_{k_ik_j}$ is the corresponding element in $\overline{\mathbf{A}}$ $\textbf{e}_{\mathit{avg}} \in \mathbb{R}^{1024}$. Then we stack another MLP $h_\phi$ on top of $\textbf{e}_{\mathit{avg}}$ to yield the prediction:
\begin{equation}\label{eq:hphi_for_cross_attn}
    \hat{y} = h_\phi(\textbf{e}_{\mathit{avg}})
\end{equation}
Subsequently, given the label of each pair $(t_i, t_j)$ as positive or negative, we utilize CrossEntropy loss to perform backward propagation and update the two MLPs in Eq.\eqref{eq:ftheta_for_cross_attn} and Eq. \eqref{eq:hphi_for_cross_attn}. Following training, we use the first MLP in Eq. \eqref{eq:ftheta_for_cross_attn} to derive the attention score $\mathbf{A}$ prior to normalization. Then we calculate the attention score of every word pair between $t_i$ and $t_j$. By averaging the attention scores across rows and columns, we identify the most significant words, $w_i$ in $t_i$ and $w_j$ in $t_j$, which inform the definition of $f_\theta$ in Cross-Attention:
\begin{equation}
    w_i, w_j = f_\theta(t_i, t_j)
\end{equation}

\section{Additional Experiments}

\subsection{Case Study}
\label{ssub:case_study}
We provide more case studies in Table \ref{tab:more_cases_overall} (Generated results of combining two stages together) and Table \ref{tab:more_cases_in_stage2} (Generated results of Stage II).

\begin{table*}[t!]
    \footnotesize
    \centering
    \begin{tabular}{c|p{13cm}}
\toprule
\textbf{Item/Feature} & \textbf{skirt dress/lemonbaby Peppa pig cartoon girls cotton birthday baby; costume/Peppa pig t ballerina } \\
\midrule
\textbf{PEPLER-F} & The peppa-colored skirt complements the pink and blue colors in the dress. the peppa-colored skirt also adds a pop of color to the outfit. \\
\midrule
\textbf{ChatGPT}  & The Lemonbaby Peppa Pig cartoon girls cotton birthday baby skirt dress and Peppa Pig T ballerina costume match because they both feature Peppa Pig as the main character and use similar color schemes and design elements. \\
\midrule
\textbf{Flan-T5-xl-F}  & The lemonbaby pig girls baby skirt dress and the peppa t ballerina costume complement each other because they both have a playful and playful vibe, which creates a cohesive outfit.\\
\midrule
\midrule
\textbf{Item/Feature} & \textbf{sock/cutieful women nylon mmhg compression dark flower power; shoe/cutieful therapeutic graduated mmhg compression dapper dogs women sizes socks} \\
\midrule
\textbf{PEPLER-F} & they are both made of the same fabric and have a similar color scheme. they complement each other because they create a cohesive outfit by balancing the proportions of the outfit. the combination of a feminine and masculine look. \\
\midrule
\textbf{ChatGPT}  & Both Cutieful Women Nylon MMHG Compression Dark Flower Power Socks and Cutieful Therapeutic Graduated MMHG Compression Dapper Dogs Women Sizes Socks provide comfortable and stylish compression support for your legs and feet. \\
\midrule
\textbf{Flan-T5-xl-F-RE}  & They complement each other because the cuteness of the graduated dapper dogs socks adds a playful and playful element to the outfit, while the compression sock provides a more serious and structured look.\\
\midrule
\midrule
\textbf{Item/Feature} & \textbf{knot/sterling silver thick love post earrings; jewelry/lifetime round ball stud earrings inch mm wide k gold semi precious metals surgical steel posts safe for most ears guaranteed life} \\
\midrule
\textbf{PEPLER-F} & the silver chain necklace and bracelets are both silver and have a polished finish that complements the silver chain necklace and bracelets. the silver chain necklace and bracelets are a great accessory for a statement piece.  \\
\midrule
\textbf{ChatGPT}  & The reason why these earrings are great is that they are made from high-quality materials such as sterling silver, semi-precious metals, and surgical steel with a guaranteed lifetime. They are also designed to match any outfit making them a great addition to any jewelry collection. \\
\midrule
\textbf{Flan-T5-xl-F-RE}  & They complement each other because the sterling knot adds a touch of elegance and sophistication to the jewelry, while the lifetime round inch wide metal creates a sleek and polished look.\\
\bottomrule
    \end{tabular}
    \caption{Generation results of combining two stages together. }
    \label{tab:more_cases_overall}
\end{table*}

\begin{table*}[t!]
    \footnotesize
    \centering
    \begin{tabular}{c|p{13cm}}
\toprule
\textbf{Items} & 'jackets', 'wool capes'  \\
\textbf{Features} & 'tailored with roundness', 'made for interesting silhouettes and brought freshness to saturated tailoring landscape' \\
\midrule
\textbf{GPT2-F} & The pairing of a tailored jacket and layered tailoring creates a unique and interesting silhouette that is complemented by the addition of freshness to the wool capes. \\
\midrule
\textbf{Flan-T5-large-F} & The roundness of the jackets complements the interesting silhouettes and wool capes add freshness to the saturated tailoring landscape. \\
\midrule
\textbf{Flan-T5-xl-F} & They complement each other because the roundness of the jackets creates an interesting silhouette, while the wool capes add a freshness to the saturated tailoring landscape.\\
\midrule
\textbf{ChatGPT} & Tailored round jackets and wool capes create unique and refreshing silhouettes, adding depth to the saturated tailoring trend. They match well together due to their similar structured styles. \\
\midrule
\textbf{Ground Truth} & The roundness of the tailored jackets and wool capes complement each other and create interesting silhouettes, adding a fresh perspective to a saturated tailoring landscape.\\
\midrule
\midrule
\textbf{Items} & 'skirt', 'boots' \\
\textbf{Features} &'frilly blush', 'made of leather, feminine touch'\\
\midrule
\textbf{GPT2-F} &  They complement each other because the feminine touch of the leather boots contrasts with the fun and playful vibe of the frilly skirt, creating an interesting and balanced outfit. \\
\midrule
\textbf{Flan-T5-large-F} & The frilly blush skirt complements the feminine touch of the leather boots, creating a balanced outfit.\\
\midrule
\textbf{Flan-T5-xl-F} & They complement each other because the frilly blush skirt adds a feminine touch to the masculine leather boots, creating a balanced outfit.\\
\midrule
\textbf{ChatGPT} & The combination of a frilly blush skirt and leather boots with a feminine touch creates a balance of softness and edginess, making for a fashionable and unique outfit. \\
\midrule
\textbf{Ground Truth} & They complement each other because the frilly blush skirt adds a hint of softness to the edginess of the leather boots, creating a balanced and harmonious outfit. \\
\midrule
\midrule
\textbf{Items} & 'shirt', 'pants' \\
\textbf{Features} & 'ruffled button-down', 'black printed cigarette'\\
\midrule
\textbf{GPT2-F} &  The ruffled button-down shirt adds a feminine touch to the outfit, while the black printed cigarette pants provide a neutral base that complements both items. \\
\midrule
\textbf{Flan-T5-large-F} & The black printed cigarette pants and ruffled button-down shirt create a cohesive outfit that is both stylish and professional. \\
\midrule
\textbf{Flan-T5-xl-F} &  The ruffled button-down shirt and black printed cigarette pants complement each other because they create a contrast in textures and patterns, creating a visually interesting outfit. \\
\midrule
\textbf{ChatGPT} & The combination of a ruffled button-down shirt and black printed cigarette pants match due to the contrast between the structured and feminine top and the sleek and edgy pants for a balanced yet bold outfit. \\
\midrule
\textbf{Ground Truth} & The ruffled button-down shirt adds texture and volume to the outfit, while the black printed cigarette pants provide a sleek and sophisticated contrast. \\
\bottomrule
    \end{tabular}
    \caption{Generation results in testset of PFE dataset of Stage II.}
    \label{tab:more_cases_in_stage2}
\end{table*}